\documentclass[letterpaper]{article}
\usepackage{spconf,amsmath,amssymb,graphicx}
\usepackage{booktabs,multirow}
\usepackage[table]{xcolor}
\usepackage{url}
\usepackage{cite}        
\usepackage{microtype}

\usepackage{enumitem}
\setlist{nosep, leftmargin=14pt}

\title{Do Saliency Maps Fail Dense Breasts? A Chance-Normalized,
Confounder-Aware Audit of Explanation Fairness in Mammography}

\name{Ahmed Jaheen}
\address{New York University, New York, NY, USA}

\begin{document}
\ninept   
\maketitle

\begin{abstract}
Saliency maps, offered as evidence that mammography classifiers use the
cancer, are untested in dense breasts, which can hide cancers. We tested
whether they locate cancers worse there, and why. On 1,345 malignant
CBIS-DDSM views, we scored three CNNs' Grad-CAM, Integrated Gradients and
occlusion maps by whether their peak hits the lesion and a
chance-normalized lesion-to-background saliency ratio (bits; $+$1 doubles
it). Unadjusted gaps were large: Grad-CAM peaks hit 44\% of dense vs.\ 58\%
of non-dense lesions. Case mix, chiefly more calcifications in dense
breasts, explained about half or more of each gap. Occlusion kept a
significant average of $-$0.56-bit gap. Controls also implicate the classifiers: unadjusted gaps
exceeded an image-only conspicuity map's, and models relied less on
dense-breast lesions. We recommend reporting density gaps before and after
case-mix adjustment, per classifier, with conspicuity and lesion-reliance
controls.
\end{abstract}


\section{Introduction}
\label{sec:intro}

A saliency map is a heat map of the image regions that most influenced a
classifier's output. Such maps are routinely shown with mammography
classifiers: a map that lights up the lesion is read as evidence that the
malignancy score rests on the cancer~\cite{pesapane2026xai}. Breast density,
the share of fibroglandular tissue, is graded in four BI-RADS (Breast
Imaging Reporting and Data System) categories, from 1 (almost entirely
fatty) to 4 (extremely dense). Dense tissue looks white on a mammogram, as
many lesions do, and can hide them (masking)~\cite{boyd2007mammographic}. Deep models
detect and classify lesions less reliably in dense
breasts~\cite{garrucho2023highdensity,woo2025subgroup}. Unlike demographic
attributes, density changes the image itself. If saliency maps also degrade
with density, the patients whose images are hardest to read would receive
the least reliable visual evidence.

Studies of saliency maps in mammography compare them with lesion
annotations only in aggregate or per finding
type~\cite{pertuz2023saliency,cerekci2024quantitative}.
Pertuz et al.~\cite{pertuz2023saliency} called for analyses by density, and
a 2026 systematic review~\cite{pesapane2026xai} found none. Group gaps in
explanation quality occur in tabular and text
data~\cite{balagopalan2022road,dai2022fairness} and can stem from the data
rather than from the saliency method~\cite{mhasawade2024understanding}. In
medical imaging, saliency maps have been compared between demographic
groups~\cite{stanley2022fairness,haupt2026explainable},
but without lesion-level ground truth, so a difference between maps cannot
be read as a difference in quality. Subgroup audits in mammography that
adjust for case differences cover classifier performance
only~\cite{woo2025subgroup}.

Two common localization scores are the pointing game (does the map's peak
fall on the lesion?) and the share of saliency inside the lesion mask, the
radiologist's outline of the lesion. A fair comparison across density must
handle two problems. First, each score has a chance level, the value that a
map with no information reaches, which depends on lesion area and
location~\cite{arun2021assessing}. A map of typical lesion locations can
outscore trained classifiers~\cite{bathineni2026metric}, and small lesions are harder to
localize~\cite{saporta2022benchmarking}. Second, dense and non-dense
breasts hold different cases. Among malignant CBIS-DDSM~\cite{lee2017curated}
views, dense breasts are smaller (median 159 vs.\ 198\,cm$^2$). Their
lesions are more often calcifications (58.7\% vs.\ 31.8\%), clusters of tiny
calcium specks, and less often masses, which are solid lumps. Such
differences in case mix, the kinds of lesions and images in each group, are
confounders: they differ between the groups and also affect the score.

\begin{figure*}[t]
  \centering
  \includegraphics[width=\textwidth]{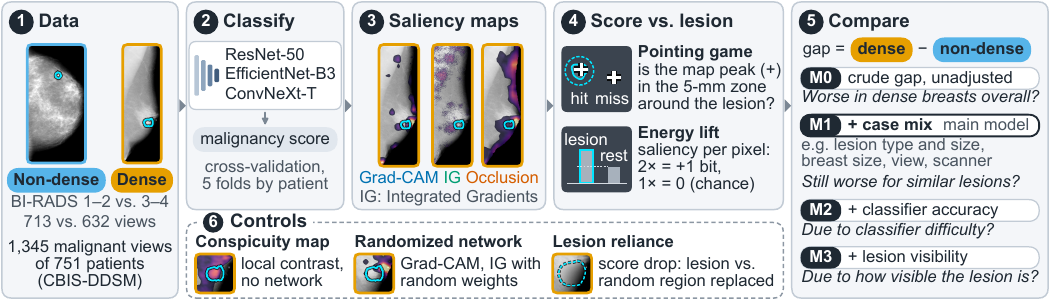}
  \caption{Study overview. Cyan: lesion (solid) and 5-mm zone (dashed).
  Images: the two mass views of Fig.~\ref{fig:examples}, selected by its
  rule; steps 3 and 6 show the dense view (orange frame), with
  network maps from ResNet-50. M0 to M3 add factors one at a time and
  re-estimate the gap (Fig.~\ref{fig:decomp}).}
  \label{fig:overview}
\end{figure*}

We test whether saliency maps are fair across density: whether, for
comparable lesions, they locate the cancer equally well, beyond chance, in
dense (BI-RADS 3 or 4) and non-dense (1 or 2) breasts. Localization alone is
a weak proxy for explanation quality~\cite{ciardiello2026beyond}, since a
map can mark the lesion while the classifier relies elsewhere. We therefore
also test whether the maps are faithful (mark the pixels the classifier
uses) and whether the classifier relies on the lesion. Fig.~\ref{fig:overview} outlines the study. Our
contributions are:
\begin{itemize}
  \item To our knowledge, the first audit of post-hoc saliency maps of
    mammography classifiers against per-lesion masks across breast density,
    for Grad-CAM, Integrated Gradients (IG) and occlusion. Maps are scored
    against chance~\cite{kohlbrenner2020towards,arun2021assessing} with the
    pointing game and the energy lift, a lesion-to-background saliency ratio
    in bits. We add a faithfulness test (ROAD~\cite{rong2022road}) and a
    lesion-reliance test.
  \item Large crude gaps for every method (Grad-CAM: 14 fewer lesion hits
    per 100 cancers). Case mix, chiefly lesion type, explains most of the
    Grad-CAM and IG energy-lift gaps and about half of occlusion's.
    Occlusion keeps a significant gap in our primary test, concentrated in
    one architecture; with method-specific case-mix adjustment
    (exploratory), no gap is significant.
  \item An audit recipe: crude and adjusted gaps with three controls. The
    crude gaps exceed a model-free conspicuity map's, vanish with randomized
    network weights (Grad-CAM; IG on ResNet-50) and come with lower lesion
    reliance, so they partly reflect what the classifiers use.
\end{itemize}

\section{Methods}
\label{sec:methods}

\subsection{Data and classifiers}
\label{ssec:data}
CBIS-DDSM~\cite{lee2017curated,sawyerlee2016cbisddsm} provides 3,568
radiologist-outlined lesions on 3,103 digitized film images of 1,566
patients. A view is one breast imaged from above (craniocaudal, CC) or from
the side (mediolateral oblique, MLO); merging 71 duplicates leaves 3,032
views. A view is malignant if any of its lesions is; its lesion mask $M$
combines its malignant outlines. The primary set $P$ holds the 1,345
malignant views with known density (751 patients). Density was graded with
the BI-RADS lexicon that preceded the 5th edition; categories
1 and 2 are
non-dense (713 views), 3 and 4 dense (632). The breast mask $B$ is the
largest region above 2\% of the intensity range, joined with $M$.

Three convolutional neural networks (CNNs), ResNet-50~\cite{he2016deep},
EfficientNet-B3~\cite{tan2019efficientnet} and
ConvNeXt-T~\cite{liu2022convnet}, start from ImageNet
weights. Each outputs a malignancy logit $z$, the
score before the sigmoid. Every view holds a lesion that a radiologist
found, so the task is to tell malignant from benign lesions. It is not
screening: cancers fully hidden by dense tissue are absent. Patient-grouped 5-fold
cross-validation gives 15 classifiers (one per architecture and fold), so
each view is classified and explained by a classifier that never saw its
patient. Training used AdamW with early stopping on the area under the
receiver operating characteristic curve (AUC) of a validation split inside
the training folds; settings were tuned on one fold. A temperature $T$,
which calibrates the scores, and a decision threshold $t$ are fitted on the
validation logits.

\subsection{Saliency maps and how we score them}
\label{ssec:metrics}
\begin{figure}[t]
  \centering
  \includegraphics[width=\columnwidth]{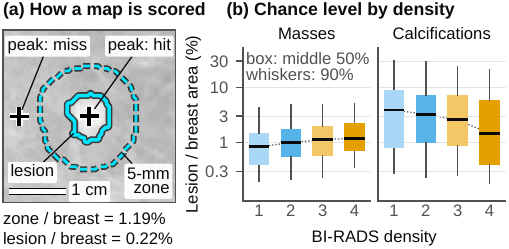}
  \caption{(a) One view (mass, density~2) with two illustrative map
  peaks. Zone/breast is the pointing-game chance level
  $\pi_{\mathrm{PG}}$; lesion/breast is the energy-lift chance level $\pi$.
  (b) $\pi$ by BI-RADS density. Calcification outlines enclose the whole
  cluster, so their $\pi$ is larger. $\pi$ varies more within groups than
  between densities, so each view is scored against its own chance level.}
  \label{fig:geom}
\end{figure}
Each map explains the logit $z$. Grad-CAM~\cite{selvaraju2017gradcam} weights the last convolutional feature
maps by their gradients, giving a coarse map with cells of about 6\,mm.
Integrated Gradients (IG)~\cite{sundararajan2017axiomatic} sums gradients
along a straight path from a black image to the view, with 64 to 512
steps, aiming for attributions that add up to the change in $z$ within 5\%
(completeness). Occlusion~\cite{zeiler2014visualizing} records how much $z$
drops when a 12.5-mm window, moved in 6-mm steps, is blurred. Only the
positive part $S\geq0$ of each map is scored, inside $B$.

The pointing game~\cite{zhang2018topdown} asks whether the peak of the map,
after 1-mm smoothing, lies within 5\,mm of the lesion, inside the zone
$D_5$ (Fig.~\ref{fig:geom}a). If so, the map scores a hit. A peak at a
random breast pixel hits with probability $\pi_{\mathrm{PG}}=|D_5|/|B|$,
the chance level. The energy lift $\lambda_E$ asks how much more saliency
per pixel the lesion gets than the rest of the breast. Let
$E_{\mathrm{in}}=\sum_M S/\sum_B S$ be the lesion's share of the saliency
and $\pi=|M|/|B|$ its share of the breast area, which is what a uniform map
would give (Fig.~\ref{fig:geom}b). Then
\begin{equation}
  \lambda_E=\log_2\frac{E'_{\mathrm{in}}/|M|}{(1-E'_{\mathrm{in}})/|B\setminus M|},\qquad
  E'_{\mathrm{in}}=\frac{E_{\mathrm{in}}+\alpha\pi}{1+\alpha}.
  \label{eq:lam}
\end{equation}
Mixing in 1\% of a uniform map
($\alpha=0.01$) keeps $\lambda_E$ finite when the lesion gets no saliency;
it then sits at its floor, about $-$6.7\,bit. A uniform map scores
$\lambda_E=0$ at any lesion size; $\lambda_E$ is therefore
chance-normalized~\cite{kohlbrenner2020towards,arun2021assessing}. Each bit
is a factor of two: at 1\,bit the lesion gets twice the background
saliency per pixel, and a gap of $-$0.5\,bit means a 29\% lower
lesion-to-background ratio. The ceiling of $\lambda_E$ still rises as
lesions shrink, so M1 to M3 adjust for lesion area.

\subsection{Comparing dense and non-dense breasts}
\label{ssec:stats}
A gap is the dense-group value minus the non-dense value; a negative gap
means worse maps in dense breasts. The crude gap is the plain difference of
the group means (views averaged over architectures). It describes the maps
that dense-breast patients receive. Its 95\% confidence interval (CI) comes
from 2,000 bootstrap resamples of patients.

To separate density from case mix, we regress each score on density and
other factors with generalized estimating equations
(GEE)~\cite{liang1986longitudinal}, a regression that gives valid CIs when
several views come from one patient. One regression, with exchangeable
correlation and bias-reduced standard errors~\cite{mancl2001covariance},
covers all architectures and methods, with one intercept per classifier and
method.
Its density coefficient $\theta$ is the adjusted gap, an association and not
a causal effect. Each factor's coefficient (slope) is shared across methods,
so all three get the same case-mix adjustment and their gaps are directly
comparable; method-specific slopes are a sensitivity analysis.
Four nested models add factors, and each answers one question
(Fig.~\ref{fig:decomp}):
\begin{itemize}
  \item M0 has density only: how much worse are maps in dense breasts?
  \item M1 adds case mix: log lesion and breast area, lesion type (mass or
    calcification), view, multiple lesions, an ill-defined margin, and an
    acquisition proxy (image stored left-right flipped, a marker of scanner
    or site).
    Among comparable lesions, does density still matter? M1 is the main
    model.
  \item M2 adds classifier accuracy on the view: whether the classifier was
    correct and how confident it was (calibrated margin $|z-t|/T$). Is the
    gap explained by how hard the view was for the classifier?
  \item M3 adds lesion visibility: the radiologist's subtlety rating and
    how much the lesion stands out in the image (the DoG map's $\lambda_E$,
    Sec.~\ref{ssec:controls}). Is the gap explained by how visible the
    lesion is?
\end{itemize}
Masking makes lesions harder to see and to classify, so the M2 and M3
factors may lie on the path from density to the map. Their gaps describe
what remains once these factors are held equal. Masking also affects which
lesions are detected, so lesion type, margin, lesion area and multiple
lesions may follow from density too, whereas acquisition, view and breast
area precede it. M1 may therefore remove part of a real density effect, and
an exploratory confounder-only model keeps only these three. For the
pointing game, a logistic GEE gives the
adjusted gap in hits per 100 cancers (percentage points, pp) by
g-computation~\cite{robins1986new}, which compares predicted hit rates with
all views set to dense and then to non-dense. We also refit M1 within
masses and within calcifications (28 mixed views excluded).

The primary test is each method's M1 $\lambda_E$ gap, Holm-corrected over
the three methods~\cite{holm1979simple}. Holm's method raises $p$-values so
that the chance of any false positive stays at most 5\%. A second Holm
family of 25 tests covers the pointing-game and faithfulness (SRG) gaps,
the density
$\times$ type interaction and the chance-level and control contrasts. A
non-significant gap is not proof of no gap, so two one-sided tests
(TOST)~\cite{schuirmann1987comparison} check equivalence: an M1 gap is
equivalent to zero if its 90\% CI lies within $\pm$0.5\,bit (a 1.41-fold
ratio) for $\lambda_E$, or $\pm$10\,pp for the pointing game. Analyses added after the main analysis are
marked exploratory. All results use one training seed, plus two more seeds
for Grad-CAM.

\subsection{Controls: image, classifier or saliency method?}
\label{ssec:controls}
The difference-of-Gaussians (DoG) conspicuity map,
$\max(0,I*g_1-I*g_8)$, subtracts the image $I$ blurred at 8\,mm from $I$
blurred at 1\,mm. It marks small spots that stand out from their
surroundings and uses no classifier, so a method's gap larger than the DoG
gap is not explained by the image alone. Randomized networks, with every
learned weight replaced by a random one~\cite{adebayo2018sanity}, test
whether a gap needs learned weights (Grad-CAM, all classifiers; IG,
ResNet-50).

Faithfulness asks whether a map marks the pixels the classifier uses.
ROAD (RemOve And Debias)~\cite{rong2022road} replaces 1\% to 20\% of the
breast pixels with values interpolated from neighbours, most- or
least-salient pixels first (the median lesion covers 1.3\%). The symmetric
relevance gain (SRG)~\cite{bluecher2024decoupling} is how much more the
calibrated score $z/T$ drops when the most-salient pixels are removed first
than when the least-salient are, averaged over five removal fractions. A faithful map has SRG above 0; an uninformative
one, 0. Lesion-specific reliance (LSR) asks whether the
classifier uses the lesion. It is the drop in $z/T$ when $D_5$ is filled in
from the surrounding tissue, minus the mean drop for up to five same-shaped regions elsewhere in $B$.
LSR above 0 means the classifier relies on the lesion more than on random
tissue.

\section{Results}
\label{sec:results}
All CIs are 95\% and uncorrected; Holm $p$-values are corrected. On all
3,032 views, held-out AUC averaged 0.79 and was lower in dense breasts for
every architecture (0.755 vs.\ 0.820). TOST could not place this gap of
$-$0.065 ($-$0.098 to $-$0.032) within $\pm$0.05.

\subsection{Crude gaps}
\label{ssec:res_crude}
\begin{table}[b]
\centering
\small
\setlength{\tabcolsep}{3pt}
\caption{Density gaps, dense minus non-dense [95\% CI]; below 0: worse in
dense breasts. Crude: group-mean difference. Adjusted: M1 with shared
case-mix slopes. LSR: lesion-specific reliance (1,276 views with a
comparison region); adjusted LSR exploratory.}
\label{tab:theta}
\begin{tabular}{@{}lrr@{}}
\toprule
 & Crude gap & Adjusted gap (M1) \\
\midrule
\multicolumn{3}{@{}l}{\emph{Energy lift $\lambda_E$ (bit)}} \\
\quad Grad-CAM  & $-$0.86 [$-$1.23, $-$0.51] & $-$0.15 [$-$0.53, 0.23] \\
\quad IG        & $-$0.62 [$-$0.80, $-$0.45] & $+$0.10 [$-$0.14, 0.35] \\
\quad Occlusion & $-$1.27 [$-$1.72, $-$0.84] & $-$0.56 [$-$1.00, $-$0.12] \\
\multicolumn{3}{@{}l}{\emph{Pointing game (hits per 100 cancers)}} \\
\quad Grad-CAM  & $-$14.1 [$-$19.3, $-$9.2] & $-$7.6 [$-$12.4, $-$2.0] \\
\quad IG        & $-$10.4 [$-$14.9, $-$5.9] & $-$3.7 [$-$8.5, 1.3] \\
\quad Occlusion & $-$12.7 [$-$17.4, $-$8.1] & $-$6.8 [$-$11.3, $-$1.9] \\
\emph{LSR (logit)} & $-$0.34 [$-$0.52, $-$0.16] & $-$0.24 [$-$0.46, $-$0.02] \\
\bottomrule
\end{tabular}
\end{table}

Before adjustment, both scores were worse in dense breasts for every
method (Table~\ref{tab:theta}). Grad-CAM peaks hit 44\% of dense-breast
lesions and 58\% of non-dense ones, 14.1 fewer hits per 100 cancers. Chance
does not explain this: a peak placed at random would hit slightly more
often in dense breasts ($+$1.2\,pp, $-$0.2 to 2.5). Grad-CAM's crude
$\lambda_E$ gap of $-$0.86\,bit means a 45\% lower lesion-to-background
ratio.

\subsection{Case mix and adjusted gaps}
\label{ssec:res_theta}
\begin{figure}[t]
  \centering
  \includegraphics[width=\columnwidth]{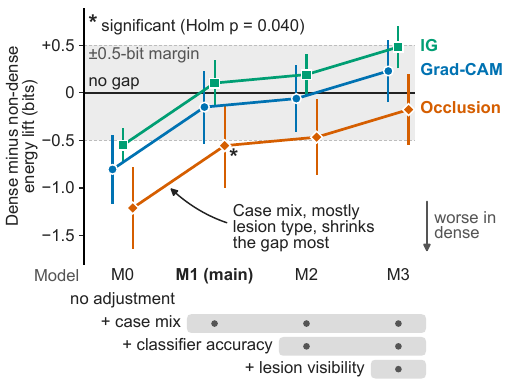}
  \caption{Energy-lift gap in each model; the rows under the axis show
  the factors each model adds. Bars: 95\% CIs; case-mix slopes shared
  across methods. With method-specific slopes (exploratory), no M1 gap is
  significant. Grey band: $\pm$0.5-bit equivalence margin, tested at M1
  with 90\% CIs. M0 is the crude gap as the regression estimates it, so it
  differs slightly from Table~\ref{tab:theta}.}
  \label{fig:decomp}
\end{figure}
\begin{figure*}[t]
  \centering
  \includegraphics[width=\textwidth]{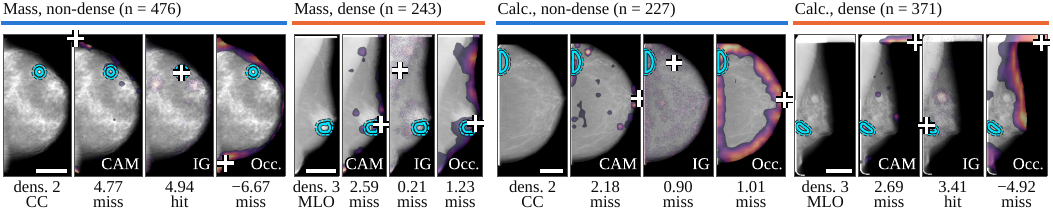}
  \caption{Rule-selected examples. For each density and lesion-type group
  ($n$ views; Calc.: calcification): the view whose ResNet-50 Grad-CAM
  $\lambda_E$ is closest to the group median, with its positive Grad-CAM
  (CAM), IG and occlusion (Occ.) maps, cropped to the breast. Cyan: lesion
  (solid) and 5-mm zone (dashed); $+$: map peak. Below each map:
  $\lambda_E$ (bits; floor about $-$6.7) and hit or miss. Bar: 5\,cm.}
  \label{fig:examples}
\end{figure*}
Adding case mix (M0 to M1, Fig.~\ref{fig:decomp}) removed 81\% of the
Grad-CAM $\lambda_E$ gap, all of IG's (118\%, leaving a slightly positive
gap) and about half of occlusion's (54\%). On the pointing game it removed
about half (45\% to 64\%). Lesion type carried 88\% of the $\lambda_E$
change (68\% to 114\%; exploratory): calcifications, more common in dense
breasts, scored 1.81\,bit lower than masses (Fig.~\ref{fig:examples}).
The exploratory confounder-only model left Grad-CAM $-$0.51\,bit ($-$0.90 to
$-$0.12), 63\% of its M0 gap ($-$0.81\,bit). Adding classifier accuracy
(M2) explained little: each $\lambda_E$ gap moved up by about 0.1\,bit.
Adding lesion visibility (M3) explained most of the rest, moving the gaps
to $-$0.18\,bit for occlusion and above 0 for the others.

In the primary test, occlusion kept a significant M1 gap of $-$0.56\,bit
(Holm $p=0.040$), a 32\% lower lesion-to-background ratio. The Grad-CAM and
IG gaps were small ($-$0.15 and $+$0.10\,bit; Holm $p=0.83$ each); their
90\% CIs lay inside $\pm$0.5\,bit, so TOST rated both equivalent to zero on
$\lambda_E$ in this model, and IG also on the pointing game. A
specification test of whether the case-mix slopes differ between methods
favoured method-specific slopes ($\chi^2_{16}=201$). With these slopes
(exploratory), the three gaps were small and did not differ between methods
($p=0.89$). They were $-$0.17\,bit for Grad-CAM ($-$0.57 to 0.23), $-$0.19
for IG ($-$0.39 to 0.02) and $-$0.24 for occlusion ($-$0.71 to 0.23). None
was significant (Holm $p$ 0.63, 0.22 and 0.63).

Occlusion's M1 gap was concentrated in EfficientNet: $-$1.19\,bit, against
$-$0.07 for ResNet-50 and $-$0.52 for ConvNeXt. Without EfficientNet it was
$-$0.27\,bit ($-$0.72 to 0.17; exploratory). EfficientNet's occlusion maps
put no positive saliency on the lesion ($\lambda_E$ at its floor) 17.3\,pp
more often in dense views (11.4 to 23.4). With two more training seeds (Grad-CAM; exploratory), the M1
gaps varied as much between seeds as between architectures (standard
deviation 0.19\,bit each), and EfficientNet's was negative in every seed
($-$0.25 to $-$0.41\,bit).

After adjustment, Grad-CAM peaks missed 7.6 more dense-breast lesions per
100 cancers and occlusion peaks 6.8 more (Table~\ref{tab:theta}). Their
CIs exclude 0, but their Holm $p$-values (second family) are 0.33 and 0.61. The remaining gaps lay in masses: there, occlusion's
M1 $\lambda_E$ gap was $-$0.72\,bit ($-$1.37 to $-$0.07), and Grad-CAM peaks
missed 12.0 more lesions per 100 (3.0 to 19.3; exploratory). No CI within
calcifications excluded 0 (density $\times$ type interaction for hits, Holm
$p=0.053$).

\subsection{Controls}
\label{ssec:res_road}
The DoG map scored 0.44\,bit lower in dense breasts, so lesions there stand
out less. Every method's crude $\lambda_E$ gap was larger than the DoG gap,
by 0.42 (Grad-CAM), 0.18 (IG) and 0.83\,bit (occlusion). All three CIs
excluded 0; after Holm correction in the second family, $p$ was 0.33, 0.67
and 0.0125. Randomized-network maps showed no density gap; the trained gaps
were larger by 0.84\,bit for Grad-CAM and 0.67\,bit for IG on ResNet-50
(both Holm $p=0.0125$). Classifiers relied less on dense-breast lesions
(lower LSR) before and after adjustment (Table~\ref{tab:theta}; adjusted
gap exploratory). SRG was above 0 in 94\%
to 99\% of views. Crude SRG gaps were negative for Grad-CAM ($-$0.20) and
occlusion ($-$0.24), but no adjusted SRG gap was Holm-significant
($p\geq0.33$).

\section{Discussion and Conclusion}
\label{sec:discussion}
In our data, dense-breast patients received worse maps, in large part
because their cancers were more often calcifications, which scored lower
than masses. Lesion type may be a confounder or, through masking and
detection, a step on the path from density to the map. If it is a
confounder, the M1 gap is the fair comparison; if it lies on that path, the
confounder-only gap is.

\emph{Limitations.} We used a single digitized-film dataset with reader-assigned
density; label noise likely biases $\theta$ toward 0, and replication on digital
mammograms is future work.

In conclusion, (i) the crude gaps were large for every method. (ii)
Accounting for case mix, chiefly lesion type, removed about half or more of
each gap; occlusion's primary-test gap was concentrated in one
architecture, and with method-specific slopes (exploratory) all three
adjusted $\lambda_E$ gaps were about $-$0.2\,bit, none significant. (iii)
The controls point partly to the classifiers: the crude gaps exceeded the
conspicuity map's, needed trained weights (Grad-CAM; IG on ResNet-50) and
came with lower lesion reliance. Audits should report crude,
confounder-only and adjusted gaps per classifier, with conspicuity,
randomized-network and lesion-reliance controls.

\section{Compliance with Ethical Standards}
\label{sec:ethics}
This research study was conducted retrospectively using de-identified human
subject data made available in open access by The Cancer Imaging Archive (CBIS-DDSM
collection~\cite{sawyerlee2016cbisddsm,lee2017curated},
CC BY 3.0). Ethical approval was not required as confirmed by the license
attached with the open access data.
    
\section{Acknowledgments}
\label{sec:acknowledgments}
The authors declare no conflicts of interest; no funding was received.
Claude was used to help in re-writing the draft. The authors verified all content and take full responsibility for it.

\let\origthebibliography\thebibliography
\renewcommand{\thebibliography}[1]{\origthebibliography{#1}\setlength{\itemsep}{0pt}\setlength{\parsep}{0pt}}
{\small
\bibliographystyle{IEEEbib}
\bibliography{refs}
}

\end{document}